\documentclass{article}
\pdfoutput=1

\PassOptionsToPackage{numbers, compress}{natbib}

\usepackage[final]{icbinb2021}

\usepackage[utf8]{inputenc} 
\usepackage[T1]{fontenc}    
\usepackage{hyperref}       
\usepackage{url}            
\usepackage{booktabs}       
\usepackage{amsfonts}       
\usepackage{nicefrac}       
\usepackage{microtype}      
\usepackage{xcolor}         

\usepackage{todonotes}
\usepackage{amsmath}
\usepackage{mathtools}
\usepackage{amssymb}
\usepackage{bbm}
\usepackage{bm}
\usepackage{wrapfig}
\usepackage{caption}
\usepackage{subcaption}
\usepackage{listings}

\title{Unbiased Gradient Estimation with Balanced Assignments for Mixtures of Experts}

\author{%
  Wouter Kool\thanks{Corresponding author: 
  \texttt{w.w.m.kool@uva.nl}. Work done on an internship at DeepMind.} \\
  University of Amsterdam\\
   \And
   Chris J. Maddison \\
   DeepMind \\
   \And
   Andriy Mnih \\
   DeepMind \\
}

\begin{document}

\maketitle

\begin{abstract}
    Training large-scale mixture of experts models efficiently on modern hardware requires assigning datapoints in a batch to different experts, each with a limited capacity. Recently proposed assignment procedures lack a probabilistic interpretation and use biased estimators for training. As an alternative, we propose two unbiased estimators based on principled stochastic assignment procedures: one that skips datapoints which exceed expert capacity, and one that samples perfectly balanced assignments using an extension of the Gumbel-Matching distribution \citep{mena2018learning}. Both estimators are unbiased, as they correct for the used sampling procedure. On a toy experiment, we find the `skip'-estimator is more effective than the balanced sampling one, and both are more robust in solving the task than biased alternatives.
\end{abstract}

\section{Introduction}
A mixture of experts (MoE) model can be used to implement conditional computation by processing different datapoints by different expert modules. This enables increasing the MoE models' representational capacity by adding more experts, without increasing the amount of computation per datapoint, which remains the same as for a single expert. Recently, this idea has been combined with deep neural networks, where each layer can be a separate MoE model, resulting in large-scale MoE's that yield state-of-the-art performance in various tasks \citep{shazeer2017outrageously,lepikhin2021gshard,fedus2021switch,lewis2021base,riquelme2021scaling,yang2021exploring}. Training an MoE model involves training a \emph{routing network} to assign datapoints to experts, and training individual experts to perform well on the datapoints assigned to them. In practice, for computational efficiency, we train on minibatches of datapoints, and as each expert has a limited capacity, we either have to \emph{skip} datapoints exceeding the capacity of the experts they are assigned to, or \emph{balance} the assignments of datapoints to different experts \citep{fedus2021switch,lewis2021base}. The effect of skipping datapoints or balancing their assignments is typically not accounted for when training the routing network. Recent work has also challenged this approach by showing that in some cases better performance can be achieved using a fixed hashing-based routing strategy \citep{roller2021hash}, which suggests that there is room for improvement in training the routing network. As a step towards this goal, we present two principled methods for optimizing MoE's under limited expert capacity. Specifically, we propose two sampling procedures and corresponding unbiased estimators in this paper: a simple one based on skipping datapoints that exceed expert capacity, and a more advanced one based on balanced datapoint assignment using an extension of the Gumbel-Matching distribution \citep{mena2018learning}. Whereas these sampling procedures ensure that each sample respects the expert capacity, we also propose to use the Sinkhorn algorithm to balance the assignment \emph{in expectation} before sampling, and we connect this procedure to the Gumbel-Matching distribution, which has such Sinkhorn balancing built-in.

In this paper, we consider a single-layer MoE model, but this can be easily generalized. Formally, the problem we consider is to predict a label $y$ for a datapoint $x$ using an MoE model consisting of $k$ individual experts $p_\theta(y|x,z)$ indexed by $z$ and selected using the routing network $p_\theta(z|x)$. At test time, we select the most probable expert $z^* = \arg\max_{z} p_\theta(z|x)$, while for training we optimize a smoothed objective obtained by taking the expectation over the routing decisions. The resulting objective, ELBO, is a variational lower bound on the marginal log-likelihood:
\begin{equation}
\label{eq:objective}
    \log p_\theta(y|x,z^*) \approx \underbrace{\mathbb{E}_{z \sim p_{\theta}(z|x)}[\log p_\theta(y|x,z)]}_{\text{ELBO}} \le \log \mathbb{E}_{z \sim p_{\theta}(z|x)}[p_\theta(y|x,z)] = \log p_\theta(y|x).
\end{equation}
We optimize this objective using minibatches of $n$ datapoints, while respecting the expert capacity by assigning at most $c = \frac{n}{k}$ datapoints to each expert (for simplicity, we assume no \emph{slack} capacity). We evaluate our estimators on a toy experiment, where we find that the simple `skip'-estimator is more effective than the one based on balanced sampling using the Gumbel-Matching distribution. This is a surprising result, as we designed balanced sampling (with importance weights) as a better alternative to wastefully skipping data, but it seems that the benefit of using all data is outweighed by the added variance due to the importance weights. We do however find that both estimators, which are based on REINFORCE \citep{glynn1990likelihood,williams1992simple}, are more robust than the biased alternatives using differentiable gating.

\section{Unbiased Estimation using Balanced Assignment}
For simplicity, we consider the problem of optimizing a general function $f(x,z)$ using gradient descent using minibatches of datapoints $\mathbf{x} = (x_1, ..., x_n)$ and expert assignments $\mathbf{z} = (z_1, ..., z_n)$. To simplify notation, we omit the dependence of $f(x,z)$ on $y$ and parameters $\theta$ (this can be added easily).
By combining importance sampling with REINFORCE \citep{glynn1990likelihood,williams1992simple}, we can sample from any joint proposal distribution $q(\mathbf{z}|\mathbf{x})$ with marginals $q(z_i|\mathbf{x})$ to estimate the gradient for a minibatch $\mathbf{x}$:
\begin{equation}
\label{eq:reinforce}
    \nabla \mathbb{E}_{\mathbf{z} \sim p_\theta(\mathbf{z}|\mathbf{x})}\left[\frac{1}{n} \sum_i f(x_i,z_i)\right] = \mathbb{E}_{\mathbf{z} \sim q(\mathbf{z}|\mathbf{x})}\left[\frac{1}{n} \sum_i \frac{p_\theta(z_i|x_i)}{q(z_i|\mathbf{x})} \nabla \log p_\theta(z_i|x_i) (f(x_i,z_i) - b) \right].
\end{equation}
Here $b$ is a baseline which can reduce the estimator variance (see Appendix \ref{app:proof_reinforce}).
Taking $q(\mathbf{z}|\mathbf{x}) = \prod_i p_\theta(z_i|x_i)$ recovers the standard `on-policy' REINFORCE  estimator.

\subsection{Skipping Datapoints as the Simple Solution}
Our `skip'-estimator respects expert capacity by sampling expert assignments independently and randomly subsampling the datapoints assigned to experts for which capacity is exceeded. If we assume a random order of the datapoints (or shuffle them first), simply skipping the last assignments per expert is equivalent to uniform subsampling. Let $\mathbf{z}$ be the vector of expert assignments, sampled independently from a proposal distribution $q(\mathbf{z}|\mathbf{x}) = \prod_i q(z_i|x_i)$. Let $n_j = \sum_i \mathbbm{1}_{\{z_i=j\}}$ be the number of datapoints assigned to expert $j$ (before subsampling) and $c = \frac{n}{k}$ be the expert capacity. Let $\boldsymbol{\delta} = (\delta_1, ..., \delta_n)$ with $\delta_i \in \{0, 1\}$ represent which datapoints are kept after per-expert subsampling. Correcting for the fact that the probability of datapoint $i$ being kept after subsampling is $\min \{n_{z_i}, c\}/n_{z_i}$, we obtain (see Appendix \ref{app:skipping}):
\begin{equation}
\label{eq:skipping_grad}
    \nabla \mathbb{E}_{\mathbf{z} \sim p_\theta(\mathbf{z}|\mathbf{x})}\left[\frac{1}{n} \sum_i f(x_i,z_i)\right]
    = \mathbb{E}_{\mathbf{z} \sim q(\mathbf{z}|\mathbf{x})}\left[\mathbb{E}_{\boldsymbol{\delta} | \mathbf{z}}\left[\frac{1}{n} \sum_i \delta_i \frac{n_{z_i}}{\min \{n_{z_i}, c\}} \cdot \frac{\nabla p_\theta(z_i|x_i)}{q(z_i|x)} f(x_i,z_i)\right]\right].
\end{equation}
Here we have omitted the baseline $b$ and used $\nabla p_\theta(z_i|x_i) = p_\theta(z_i|x_i) \nabla \log p_\theta(z_i|x_i)$ for brevity. Note that, while we skip datapoints in the gradient estimate \eqref{eq:skipping_grad}, we can still propagate them to subsequent layers, e.g.\ by using skip connections or \emph{no-token-left-behind} routing \citep{fedus2021switch}. In paricular, we can also apply \eqref{eq:skipping_grad} to multilayer MoE's, where we can skip different datapoints in different layers.

\subsection{Balanced Assignment using Gumbel-Matching}
\label{sec:gumbel_matching}
As an alternative to skipping, we use the $n \times k$ Gumbel-Matching distribution, a strict generalization of the ($n \times n$) Gumbel-Matching distribution \citep{mena2018learning}, to sample perfectly balanced assignments. We derive it by using the Gumbel-Max trick \citep{maddison2014sampling,gumbel1954statistical} to view sampling of independent expert assignments as an optimization problem, and adding constraints to this problem to respect the expert capacity. 
Let $z_{ij} = \mathbbm{1}_{\{z_i=j\}}$ be the one-hot representation of $z_i$, $a_{ij}$ the unnormalized log-probability (logit) of assigning datapoint $i$ to expert $j$, and $g_{ij} \sim \text{Gumbel}(0)$ i.i.d.\ standard Gumbel variables. 
Then $\mathbf{z}$ has the Gumbel-Matching distribution if it is the solution to the following optimization problem:
\begin{align}
\label{eq:gumbel_matching_problem}
    \max_{\mathbf{z}} \sum_{ij} z_{ij} ( a_{ij} / \tau + g_{ij}) \quad \quad
    \text{s.t. } \sum_{j} z_{ij} = 1 \; \forall i, \quad \sum_{i} z_{ij} \le c \; \forall j, \quad z_{ij} \in \{0, 1\} \; \forall i,j,
\end{align}
where $\tau$ is a \emph{temperature} parameter and $c = \frac{n}{k}$ is the expert capacity. If we remove the balancing/capacity constraint $\sum_{i} z_{ij} \le c$, the solution decomposes over $i$ and is given by $z_i = \arg\max_j (a_{ij} / \tau + g_{ij})$ which is equivalent to $z_i \sim \text{Categorical}(\exp(a_{ij}/\tau)/\sum_j \exp(a_{ij}/\tau))$ as a result of the Gumbel-Max trick. Thus, adding the constraint can be seen as a natural way of enforcing a balanced assignment to the otherwise independent sampling procedure.
Generalizing the result from \citep{mena2018learning}, the $n \times k$ Gumbel-Matching approximates the Gibbs distribution over $n \times k$ assignments $\mathbf{z}$ with potentials given by $\sum_{ij} z_{ij} a_{ij}$ and temperature $\tau$ (see Appendix \ref{app:gm_approximates_gibbs}):
\begin{equation}
\label{eq:gibbs_distribution}
    p(\mathbf{z}) \propto \exp\left(\frac{1}{\tau} \textstyle\sum_{ij} z_{ij} a_{ij}\right).
\end{equation}
For $\tau \to 0$ we obtain the deterministic assignment used in BASE layers \citep{lewis2021base}. We propose to solve the $n \times k$ assignment problem using a special cycle cancelling algorithm \citep{klein1967primal} (see Appendix \ref{app:gm_cycle_cancelling}), which for $k \ll n$ is more efficient than $O(n^3)$ assignment using the Hungarian Algorithm \citep{kuhn1955hungarian}. 

The marginals $q_\theta(z_i|\mathbf{x})$ of the Gumbel-Matching distribution are intractable \citep{mena2018learning}, but we can compute the \emph{conditionals} $q_\theta(z_i|\mathbf{x},\mathbf{G}_{-i})$, conditioned on the noise $\mathbf{G_{-i}} = (\mathbf{g}_1, ..., \mathbf{g}_{i-1}, \mathbf{g}_{i+1}, ..., \mathbf{g}_n)$ of the other datapoints. These can be computed efficiently (see Appendix \ref{app:gm_compute_conditionals}) and used as a stochastic approximations to the marginals, which still yield unbiased gradients when used in \eqref{eq:reinforce}. Formally, let $\mathbf{z} = \text{GM}(\log p(\cdot|\mathbf{x}),\mathbf{G})$ be the solution for the Gumbel-Matching problem with noise $\mathbf{G}$. We can then use the following estimator (see Appendix \ref{app:gm_estimator}):
\begin{equation}
\label{eq:gumbel_matching_estimator}
    \nabla \mathbb{E}_{\mathbf{z} \sim p_\theta(\mathbf{z}|\mathbf{x})}\left[\frac{1}{n} \sum_i f(x_i,z_i)\right] = \mathbb{E}_{\mathbf{G}}\left[\frac{1}{n} \sum_i \frac{\nabla p_\theta(z_i|x_i)}{q_\theta(z_i|\mathbf{x},\mathbf{G}_{-i})} f(x_i,z_i) \middle|_{\mathbf{z} = \text{GM}(\log p(\cdot|\mathbf{x}),\mathbf{G})}\right].
\end{equation}

\subsection{Bonus: Balancing Expectations using the Sinkhorn Algorithm}
\begin{wrapfigure}{r}{0.3\textwidth}
\vskip 2em
  \begin{center}
    \includegraphics[trim=8.7cm 1.5cm 8.7cm 3cm,width=0.28\textwidth]{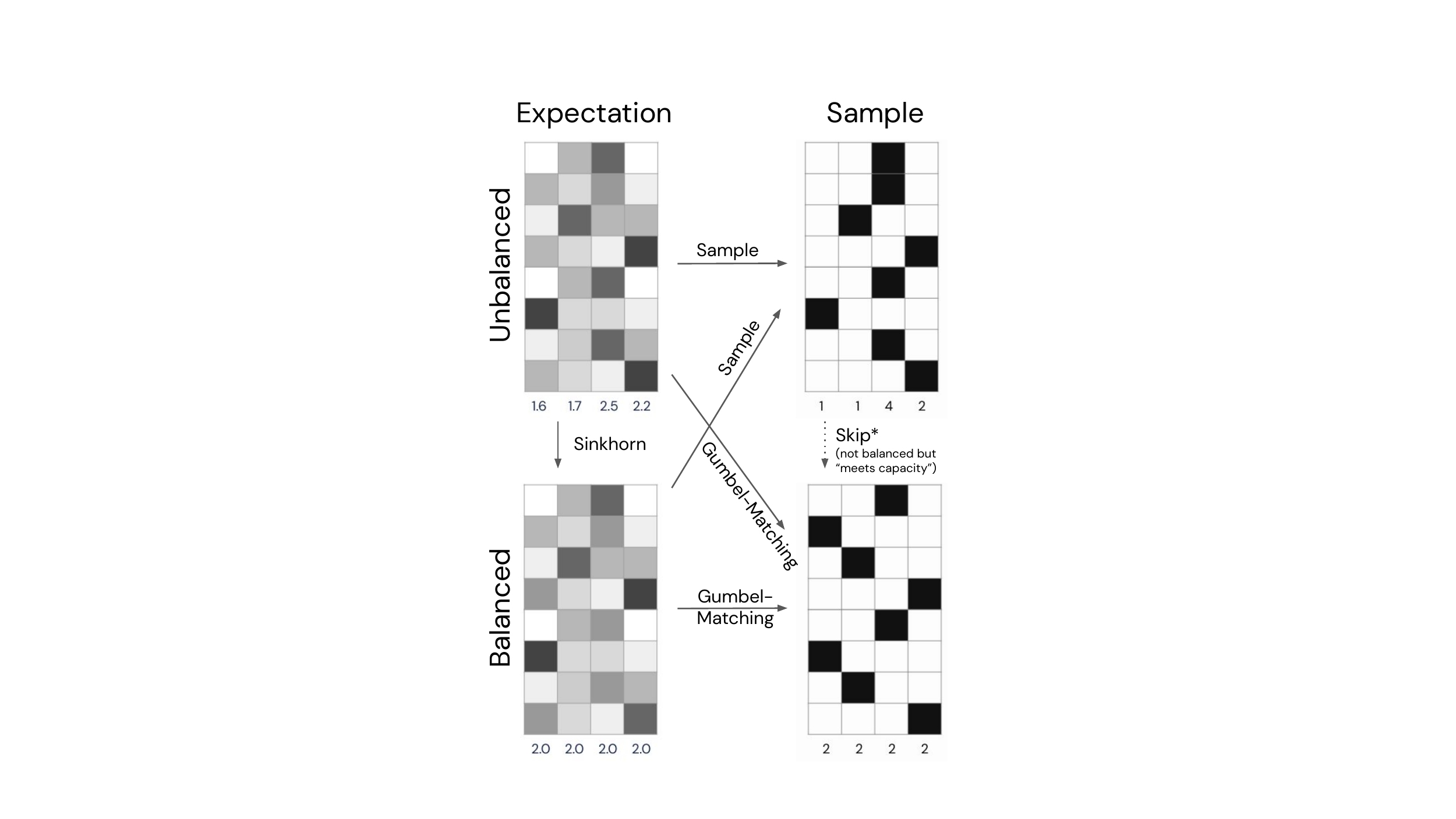}
  \end{center}
  \caption{Overview of the methods that generate balanced and unbalanced samples and expectations.}
  \label{fig:sampling_overview}
\end{wrapfigure}
\label{sec:sinkhorn_balancing}
As an alternative to generating balanced samples, we can balance the distribution \emph{in expectation} by using a generalization of the Sinkhorn algorithm \citep{sinkhorn1964relationship,sinkhorn1967concerning,knight2008sinkhorn} to non-square matrices, which iteratively normalizes the columns and rows of the probability matrix $p_{ij} = p_\theta(z_i=j|x_i)$ to sum to $\frac{n}{k}$ and $1$ respectively, until convergence. We may then use this Sinkhorn-normalized distribution as the proposal $\mathbf{q}(\mathbf{z}|\mathbf{x})$ in \eqref{eq:reinforce} and \eqref{eq:skipping_grad}, which yields more (but not perfectly) balanced samples. The Gumbel-Matching distribution is invariant to this normalization as it does not change the solution to \eqref{eq:gumbel_matching_problem}. See Figure \ref{fig:sampling_overview} for an overview.

The Sinkhorn algorithm is a direct extension of the softmax function, which solves a soft (entropy-regularized) version of the assignment problem \citep{cuturi2013sinkhorn,mena2017sinkhorn,mena2018learning}. As a result, it can be seen as approximating the marginals for the Gibbs distribution \eqref{eq:gibbs_distribution} \citep{globerson2007approximate,mena2018learning} (for $n = k$, but this can be generalized to $k < n$). Empirically, we find that the Sinkhorn algorithm (as a `soft' matching algorithm) also closely approximates the marginals of the Gumbel-Matching distribution (itself an approximation of \eqref{eq:gibbs_distribution}), at least for $n \ge 4$. As such, we propose to heuristically use it with \eqref{eq:reinforce} as an alternative to \eqref{eq:gumbel_matching_estimator}.

The Sinkhorn algorithm yields a \emph{balanced row-stochastic matrix} with probabilities/expectations $\tilde{p}_{ij}$, that can be seen as a balanced approximation of the unbalanced probabilities $p_{ij}$.
Given such a balanced matrix, there exist many distributions over balanced assignments that have the (per datapoint) marginals equal to $\tilde{p}_{ij}$, which follows from generalizing the Birkhoff theorem \citep{birkhoff1946tres,von1953certain}. When using such a distribution with stochastic gradient descent, we would like to minimize the dependence between samples for different datapoints, to reduce the variance of the gradient estimates. This can be achieved by maximizing the entropy of the joint distribution over expert assignments with the given marginals $\tilde{p}_{ij}$. In Appendix \ref{app:maxent_distribution}, we show that this maximum-entropy distribution has the form \eqref{eq:gibbs_distribution}, again motivating the Gumbel-Matching distribution as a principled approximation.

\section{Experiment}
\begin{figure}
     \centering
     \begin{subfigure}[b]{0.32\textwidth}
         \centering
         \includegraphics[width=\textwidth]{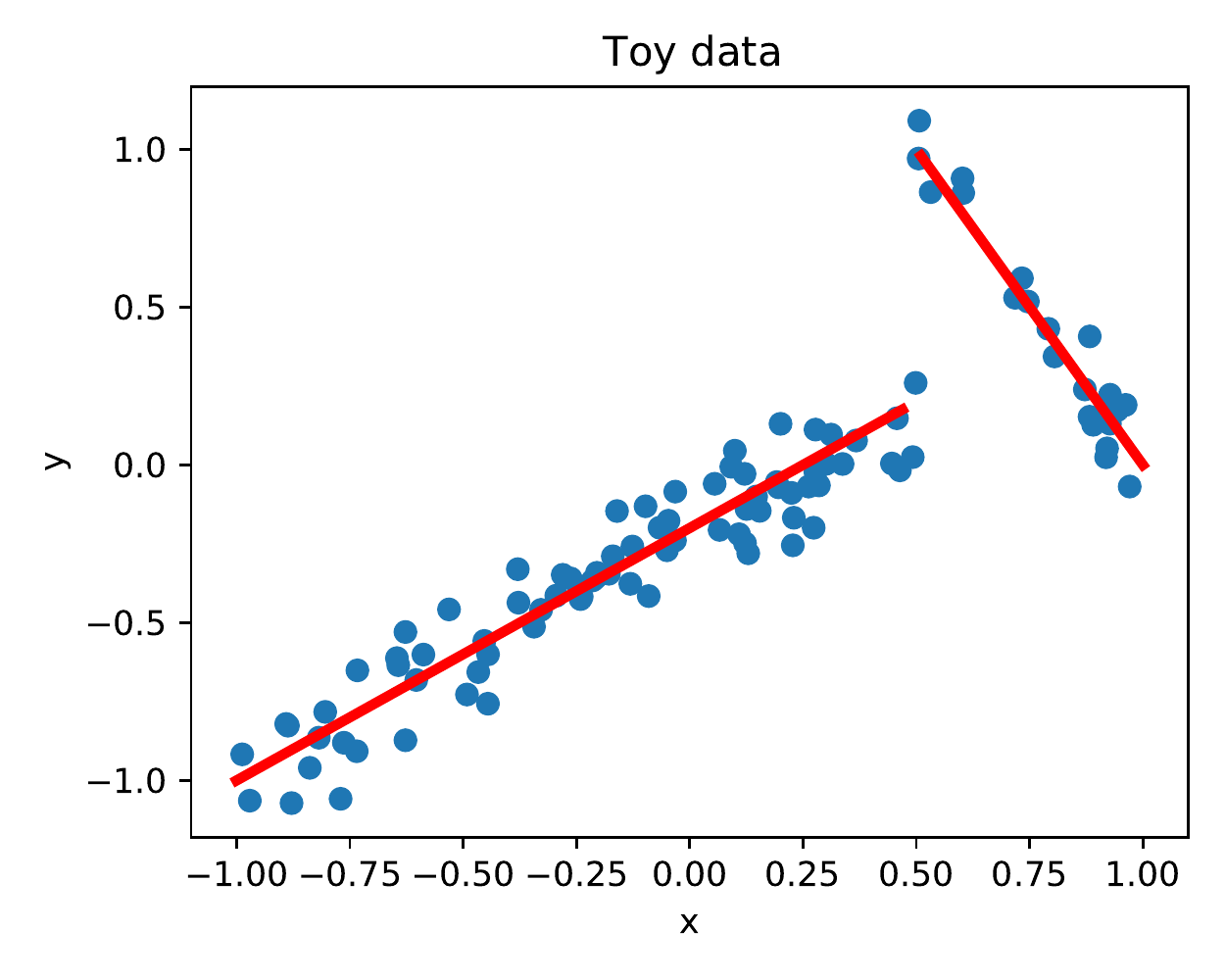}
         \caption{Toy dataset}
         \label{fig:toy_data}
     \end{subfigure}
     \hfill
     \begin{subfigure}[b]{0.32\textwidth}
         \centering
         \includegraphics[width=\textwidth]{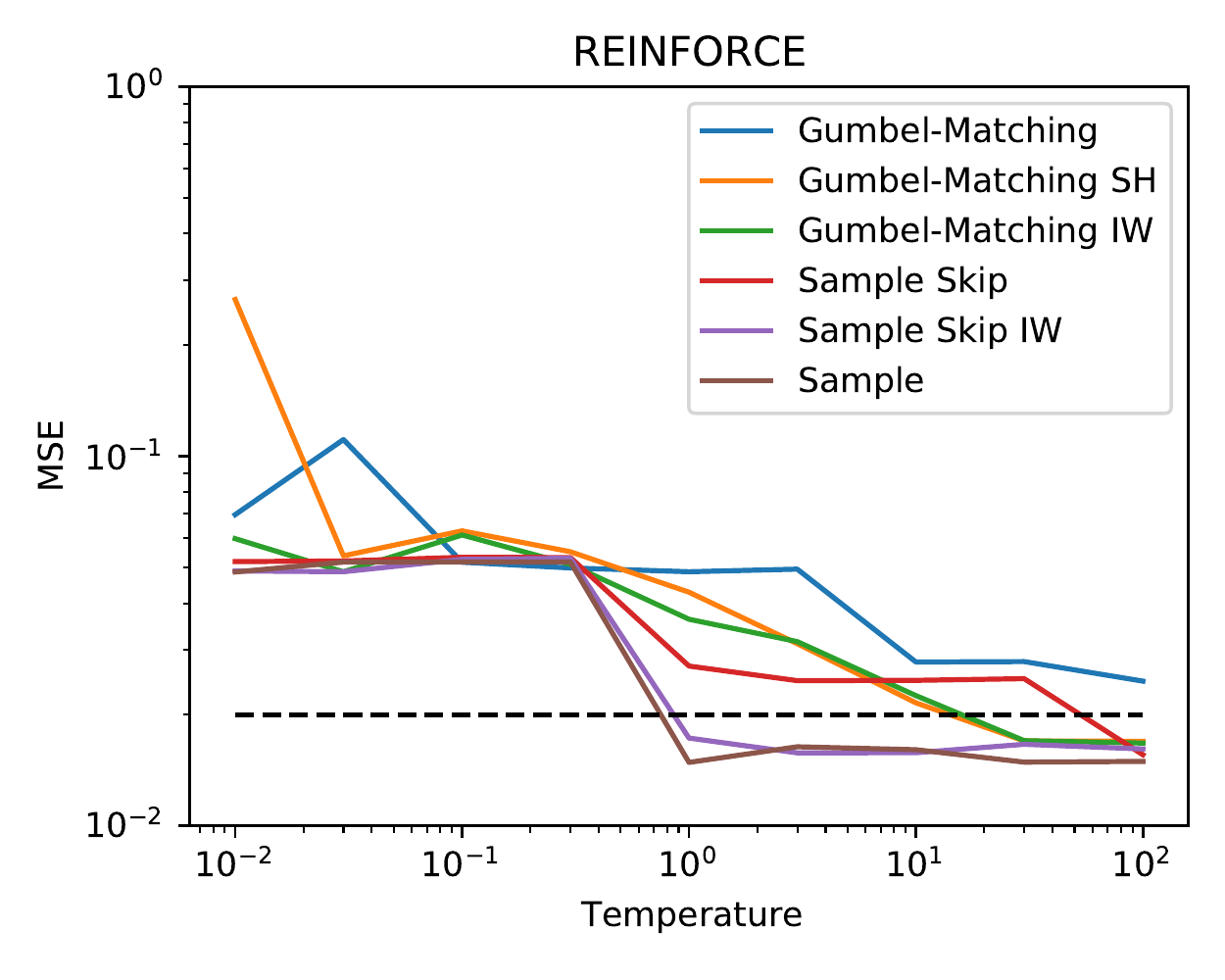}
         \caption{Results using REINFORCE}
         \label{fig:toy_results_reinforce}
     \end{subfigure}
     \hfill
     \begin{subfigure}[b]{0.32\textwidth}
         \centering
         \includegraphics[width=\textwidth]{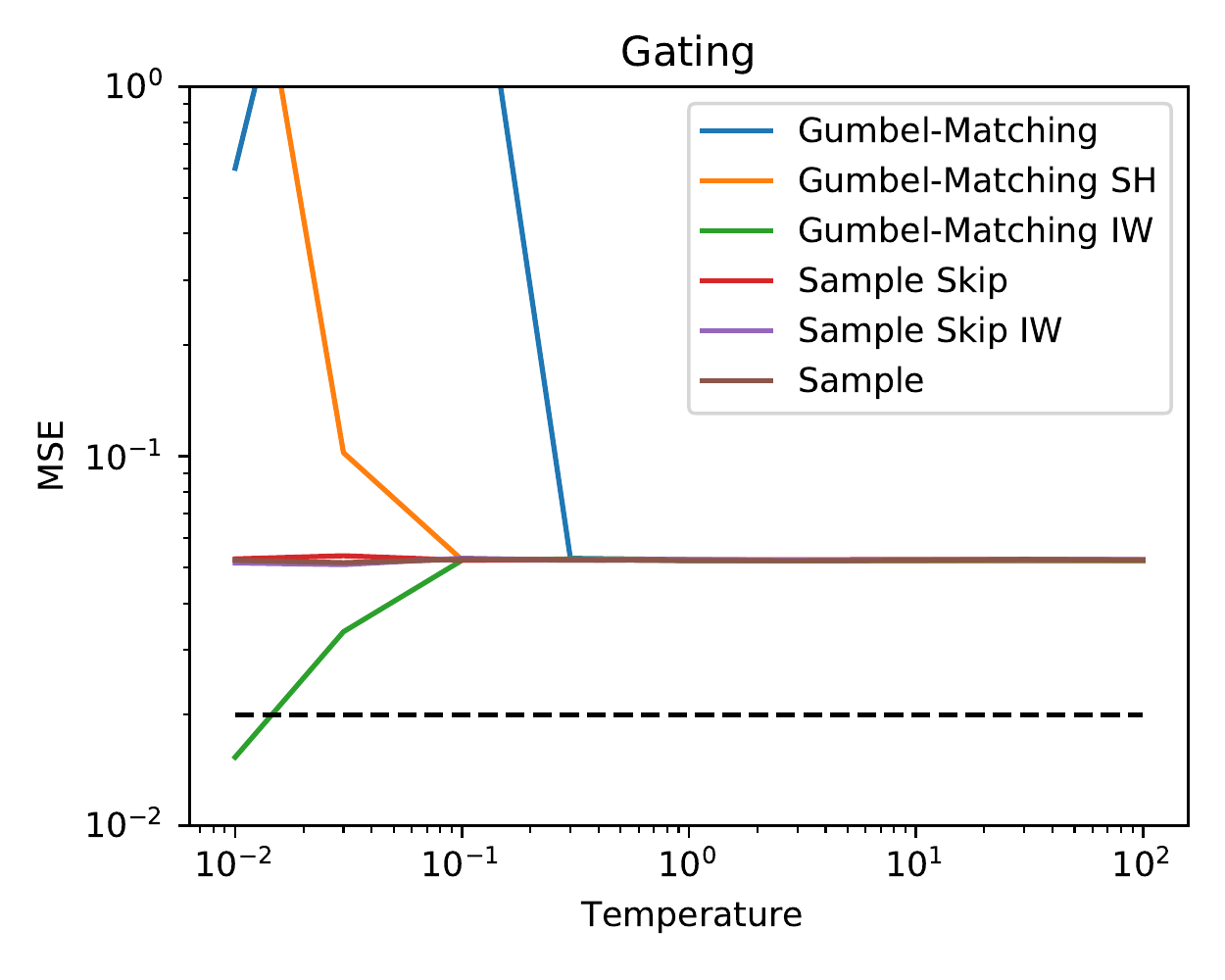}
         \caption{Results using gating}
         \label{fig:toy_results_gating}
     \end{subfigure}
     \caption{Toy experiment data and results using REINFORCE (without balance loss) and training using gating (with balance loss weight 0.01). We report final mean squared error (MSE) when training with different temperatures/noise scales. The black dashed line indicates the success threshold 0.02.}
\end{figure}

We evaluate the proposed estimators on a toy task of modelling a discontinuous function $y = \mathbbm{1}_{x<0.5} (0.8 x -0.2) + \mathbbm{1}_{x \ge 0.5} (-2.0 x + 2.0)$. We sample a dataset of 100 datapoints $x \sim \text{Uniform}(-1, 1)$ and add noise $\epsilon \sim \text{Normal}(0, 0.1^2)$, see Figure \ref{fig:toy_data}. We model this dataset using a mixture of two linear experts and a Bernoulli router distribution with probability given by the sigmoid of a linear function. This model is able to solve the task perfectly, but the training methods need to take into account the fact that the solution involves an imbalanced partitioning of the dataset between the experts. We train the model to minimize the mean squared error (MSE) under sampling of the experts, which corresponds to the objective \eqref{eq:objective} where $p_\theta(y|x,z)$ is Gaussian with a fixed variance. We consider the task solved successfully if the $\text{MSE} < 0.02$ after 10K steps of training using Adam \citep{kingma2014adam} with the learning rate of $0.1$ (found to work best overall using a grid search).

We compare the following sampling strategies combined with REINFORCE: \textbf{Sample Skip IW} is the skipping estimator with the importance-weighting correction (Equation \eqref{eq:skipping_grad}), whereas \textbf{Sample Skip} is the biased alternative that simply averages the gradient over the remaining datapoints. \textbf{Gumbel-Matching IW} uses the conditional distributions for the importance weights (Equation \eqref{eq:gumbel_matching_estimator}), whereas \textbf{Gumbel-Matching SH} is the (biased) version that uses the `Sinkhorn marginals' (see Section \ref{sec:sinkhorn_balancing}) with Equation \eqref{eq:reinforce}, and \textbf{Gumbel-Matching} is the biased estimator that does not correct for balancing using importance weights. Lastly, to quantify the reduction in performance due to the expert capacity constraints, we also include the results for \textbf{Sample}, the ideal baseline that does \emph{not} respect expert capacity. We run each experiment with 10 seeds, for a range of sampling temperatures $\tau$, taking into account their effect on the proposal distribution in the importance weights for all estimators. To reduce variance, we include an exponential moving average baseline \citep{sutton2018reinforcement} with decay 0.99.

Figure \ref{fig:toy_results_reinforce} plots the final training MSE for different temperatures $\tau$ of the sampling (proposal) distribution, where we observe that \textbf{Sample Skip IW} is the only estimator that matches the imbalanced (unconstrained) \textbf{Sample} estimator, both solving the task for $\tau \ge 1$. The skipping estimator thus provides a simple and effective way to deal with limited expert capacity, but it is important to upweight the remaining samples for experts which have skipped datapoints, as \textbf{Sample Skip}, which does not use this weighting, does not achieve the same performance. Confounding our expectation, the Gumbel-Matching based estimators turned out to be less effective, because of the increased variance due to the importance weights. Investigating the issue, we found that a datapoint $x$ can have high probability $p_\theta(z|x)$ for an expert $z$ \emph{according to the router}, but a low probability under the proposal $q(z|x)$ of \emph{actually being assigned} to the expert $z$, due to the balancing constraint. While the latter probability is small, occasionally the datapoint will get assigned to $z$, resulting in a large importance weight $\frac{p_\theta(z|x)}{q(z|x)}$. This effect can be mitigated by increasing the temperature of the proposal distribution, making it more uniform and avoiding large importance weights, which explains the good results for large $\tau$ values for all estimators except \textbf{Gumbel-Matching}. We also experiment with all estimators with Sinkhorn balancing (Section \ref{sec:sinkhorn_balancing}) before sampling, which only works for high temperatures (see Appendix \ref{app:toy_extra_results}).

\subsection{Biased training using differentiable gating}
\label{sec:experiment_gating}
Large scale MoE's used in practice \citep{shazeer2017outrageously,lepikhin2021gshard,fedus2021switch,lewis2021base,riquelme2021scaling,yang2021exploring} do not use REINFORCE, but instead multiply the output of an expert by the router probability $p_\theta(z|x)$, which we refer to as \emph{differentiable gating}. This way, the router becomes more coupled with the experts and gets a gradient signal directly from the objective. Different strategies for injecting noise to encourage exploration have been proposed, e.g.~perturbing router inputs or (log-)probabilities with multiplicative or additive (Gaussian) noise \citep{shazeer2017outrageously,fedus2021switch}. Empirically, we find that we get similar results by perturbing log-probabilities with (scaled) Gumbel noise, which, since $\arg\max_j (a_{ij} + \tau \cdot g_{ij}) = \arg\max_j (a_{ij} / \tau + g_{ij})$, has the advantage of being interpretable as sampling from a categorical distribution with the temperature $\tau$ (see Section \ref{sec:gumbel_matching}). We find training using differentiable gating succeeds only if we additionally include a load balancing loss \citep{fedus2021switch} with a weight of $0.01$ or $0.03$, and use balanced sampling with importance weights in a low temperature regime, as can be seen in Figure \ref{fig:toy_results_gating}. With differentiable gating, we may wonder if we need importance weights at all, since existing methods do not use importance weights to correct for the sampling temperature (noise scale) $\tau$. In Appendix \ref{app:toy_extra_results} we show however, that with differentiable gating, we fail to train the model when not using importance weights.

\section{Discussion}
In this paper we proposed several new estimators for training MoE models with a limited computational capacity per expert. We expected balanced sampling with importance weighting to correct for assignment of datapoints to low-probability experts to perform at least as well as skipping, which effectively uses a weight of 0 for the skipped datapoints. We found this not to be the case in practice, with the added variance from importance weights eliminating the benefit from the increased expert capacity utilization due to balanced sampling. Fortunately, the skipping estimator turned out to be a simple and effective alternative. We hope this work will be useful for training MoE models in practice.

\begin{ack}
We would like to thank Michalis Titsias, J\"{o}rg Bornschein, Matthias Bauer and Yee Whye Teh for helpful comments, discussions and support.
\end{ack}


\bibliographystyle{plain}
\bibliography{references}


\clearpage
\appendix

\section{Related work}
Mixtures of experts \citep{eigen2013learning,shazeer2017outrageously} have a long history as a method for conditional computation \citep{bengio2013estimating,bengio2015conditional,graves2016adaptive} where different experts are used for different datapoints. Typically, the \emph{router} or \emph{gating} network that assigns datapoints to experts is learned jointly with the experts themselves. Mixtures of experts and the more general \emph{routing} or \emph{modular networks} \citep{rosenbaum2019routing,kirsch2018modular,ramachandran2019diversity}, sometimes treat expert assignments as latent variables and use EM or variational methods \citep{kirsch2018modular} for training them. The benefit of optimizing a single-sample bound (ELBO) rather than marginal likelihood, is that in additional to being more tractable, it can help to avoid stochasticity  \citep{raiko2015techniques} in  datapoint assignments to experts.

Whereas the \emph{conditional} distribution over experts given a datapoint should ideally have low entropy, the \emph{marginal} distribution over experts should be balanced for efficient training and use of model capacity. In some cases this is explicitly encouraged by using a load balancing loss \citep{bengio2015conditional,fedus2021switch}. Other methods algorithmically balance the assignment \citep{lewis2021base} or use a fixed distribution \citep{roller2021hash}. Many recent large scale MoE models use heuristics to train the router module (see Section \ref{sec:experiment_gating}) but have yielded state-of-the-art performance in different domains \citep{shazeer2017outrageously,lepikhin2021gshard,fedus2021switch,lewis2021base,riquelme2021scaling,yang2021exploring}.

Sampling and optimization of balanced assignments is closely related to the sampling and optimization of permutations ($n \times n$ assignments) \citep{le2007direct,adams2011ranking,li2013efficient,mena2017sinkhorn,mena2018learning,grover2019stochastic,patrini2020sinkhorn}, which relate to estimation of the matrix permanent \citep{huber2006exact,jerrum2004polynomial,kuck2019approximating}. In a different setting, the problem can also be seen as random fair assignment \citep{budish2013designing}, with the difference being that in random fair assignment one typically is not concerned about dependence between the individual assignments.

\section{Derivation of (off-policy) REINFORCE}
\label{app:proof_reinforce}
First, we derive (off-policy) REINFORCE for a single datapoint $x$:
\begin{align*}
    \nabla \mathbb{E}_{z \sim p_\theta(z|x)}[f(x,z)] &= 
    \nabla \sum_z p_\theta(z|x) f(x,z) \\
    &= 
 \sum_z \nabla p_\theta(z|x) f(x,z) \\
    &= 
     \sum_z q(z|x) \frac{\nabla p_\theta(z|x)}{q(z|x)} f(x,z) \\
    &= \mathbb{E}_{z \sim q(z|x)}\left[\frac{\nabla p_\theta(z|x)}{q(z|x)} f(x,z)\right] \\
    &= \mathbb{E}_{z \sim q(z|x)}\left[\frac{p_\theta(z|x)}{q(z|x)} \nabla \log p_\theta(z|x) f(x,z)\right]
\end{align*}
Now consider a minibatch $\mathbf{x} = (x_1, ..., x_n)$ and let $p_\theta(\mathbf{z}|\mathbf{x}) = \prod_i p_\theta(z_i|x_i)$ be the distribution that samples expert assignments independently. Let $q(z_i|\mathbf{x}) = \sum_{z_{-i}} q(\mathbf{z}|\mathbf{x})$ be the marginal of any joint proposal distribution $q(\mathbf{z}|\mathbf{x})$, which we can use to estimate the minibatch gradient:
\begin{align*}
    \nabla \mathbb{E}_{\mathbf{z} \sim p_\theta(\mathbf{z}|\mathbf{x})}\left[\frac{1}{n} \sum_i f(x_i,z_i)\right]
    &= \frac{1}{n} \sum_i \nabla \mathbb{E}_{z_i \sim p_\theta(z_i|x_i)}\left[f(x_i,z_i)\right] \\
    &= \frac{1}{n} \sum_i  \mathbb{E}_{z_i \sim q(z_i|\mathbf{x})}\left[\frac{\nabla p_\theta(z_i|x_i)}{q(z_i|\mathbf{x})} f(x_i,z_i)\right] \\
    &= \frac{1}{n} \sum_i \mathbb{E}_{\mathbf{z} \sim q(\mathbf{z}|\mathbf{x})}\left[\frac{\nabla p_\theta(z_i|x_i)}{q(z_i|\mathbf{x})} f(x_i,z_i)\right] \\
    &= \mathbb{E}_{\mathbf{z} \sim q(\mathbf{z}|\mathbf{x})}\left[\frac{1}{n} \sum_i \frac{\nabla p_\theta(z_i|x_i)}{q(z_i|\mathbf{x})} f(x_i,z_i)\right] \\
    &= \mathbb{E}_{\mathbf{z} \sim q(\mathbf{z}|\mathbf{x})}\left[\frac{1}{n} \sum_i \frac{p_\theta(z_i|x_i)}{q(z_i|\mathbf{x})} \nabla \log p_\theta(z_i|x_i) f(x_i,z_i)\right].
\end{align*}
Since $\nabla \mathbb{E}_{z \sim p_\theta(z|x)}[b] = 0$, we can subtract any constant baseline $b$ from $f(x,y)$, resulting in \eqref{eq:reinforce}. 

\section{Unbiasedness of the skipping estimator}
\label{app:skipping}
First, consider a function $h(x_i,z_i)$. Let $z_{ij} = \mathbbm{1}_{\{z_i=j\}}$ be the one-hot representation of $z_i$ and let $h_{ij} = h(x_i,z_i)|_{z_i=j}$. Let $n_j = \sum_i z_{ij}$ be the number of datapoints assigned to expert $j$ (before subsampling). Now let $\delta_i \in \{0, 1\}$ represent which datapoints are kept after we, for each expert $j$, uniformly subsample $\min \{n_j, c\}$ datapoints, where $c = \frac{n}{k}$ is the expert capacity. If $z_{ij} = 1$ (before subsampling), then the probability that datapoint $i$ remains after subsampling is $\frac{\min \{n_j, c\}}{n_j}$, so we have $\mathbb{E}_{\boldsymbol{\delta}^* | \mathbf{z}}\left[\delta_i\right] \frac{n_j}{\min \{n_j, c\}} z_{ij} = z_{ij}$.
\begin{align*}
    \mathbb{E}_{\mathbf{z} \sim p_\theta(\mathbf{z}|\mathbf{x})}\left[\frac{1}{n} \sum_i h(x_i,z_i)\right] &= \mathbb{E}_{\mathbf{z} \sim p_\theta(\mathbf{z}|\mathbf{x})}\left[\frac{1}{n} \sum_i \sum_j z_{ij} h_{ij}\right] \\
    &= \mathbb{E}_{\mathbf{z} \sim p_\theta(\mathbf{z}|\mathbf{x})}\left[\frac{1}{n} \sum_i \sum_j \mathbb{E}_{\boldsymbol{\delta} | \mathbf{z}}\left[\delta_i\right] \frac{n_j}{\min \{n_j, c\}} z_{ij} h_{ij}\right] \\
    &= \mathbb{E}_{\mathbf{z} \sim p_\theta(\mathbf{z}|\mathbf{x})}\left[\mathbb{E}_{\boldsymbol{\delta} | \mathbf{z}}\left[\frac{1}{n} \sum_i \delta_i \frac{n_{z_i}}{\min \{n_{z_i}, c\}} h(x_i,z_i)\right]\right]
\end{align*}
Now substituting $h(x_i,z_i) = \frac{\nabla p_\theta(z_i|x_i)}{q(z_i|x)} f(x_i,z_i)$ and combining with Equation \eqref{eq:reinforce} results in \eqref{eq:skipping_grad}.

\section{The Gumbel-Matching distribution}

\subsection{Approximation to the Gibbs distribution}
\label{app:gm_approximates_gibbs}
Here we essentially reproduce the argument from \citep{mena2018learning} for the $n \times k$ Gumbel-Matching distribution. By sampling i.i.d. Gumbel noise $g_{\mathbf{z}}$ for every assignment $\mathbf{z}$, we can sample from \eqref{eq:gibbs_distribution} by maximizing
\begin{equation*}
    \left(\frac{1}{\tau} \textstyle\sum_{ij} z_{ij} a_{ij}\right) + g_{\mathbf{z}}
\end{equation*}
subject to the constraints given by \eqref{eq:gumbel_matching_problem}. Comparing this to the objective for the Gumbel-Matching problem \eqref{eq:gumbel_matching_problem}:
\begin{equation*}
    \sum_{ij} z_{ij} ( a_{ij} / \tau + g_{ij}) = \left(\frac{1}{\tau} \textstyle\sum_{ij} z_{ij} a_{ij}\right) + \sum_{ij} z_{ij} g_{ij}
\end{equation*}
we observe how the Gumbel-Matching distribution approximates \eqref{eq:gibbs_distribution} through the use of rank-one perturbations \citep{papandreou2011perturb,hazan2013sampling,tomczak2016some} $\sum_{ij} z_{ij} g_{ij}$ instead of $g_{\mathbf{z}}$.

\subsection{Solving \texorpdfstring{$n \times k$}{n times k} matching using cycle cancelling with Floyd-Warshall}
\label{app:gm_cycle_cancelling}
The $n \times k$ assignment problem can be modelled as a minimum cost flow problem which can be solved using cycle cancelling \citep{klein1967primal} as follows:
\begin{itemize}
    \item Find an initial (heuristic) feasible assignment $\mathbf{z}$. We use the auction algorithm used in \citep{lewis2021base} with $\epsilon = 1.0$ such that it finds a good (but suboptimal) solution quickly.
    \item For every combination of experts $j, j'$, find the lowest cost $d_{jj'}$ to move a datapoint from expert $j$ to $j'$. Let $s_{ij} = a_{ij} / \tau + g_{ij}$ be the \emph{score} for assigning datapoint $i$ to $j$, such that moving datapoint $i$ from $j$ to $j'$ we lose $s_{ij}$ but gain $s_{ij'}$, incurring a net `cost' of $s_{ij} - s_{ij'}$. Therefore, the minimum cost to move any of the currently assigned datapoints from $j$ to $j'$ is $d_{jj'} = \min_{i: z_{ij} = 1} s_{ij} - s_{ij'}$.
    \item Use the Floyd-Warshall algorithm\footnote{We use the parallel version: \url{https://en.wikipedia.org/wiki/Parallel_all-pairs_shortest_path_algorithm\#Floyd_algorithm} which runs in $O(k^3)$ and $k$ sequential steps.} \citep{floyd1962algorithm} to find all indirect shortest paths in the fully connected graph with $k$ nodes (one per expert) and distance from $j$ to $j'$ given by $d_{jj'}$. Stop as soon as a negative cycle is found (distance from $j$ to $j$ smaller than 0).
    \item If no negative cycle exists, the assignment is optimal, stop.
    \item For each edge $(j, j')$ in the negative cycle,\footnote{Can be reconstructed by using Floyd-Warshall with path reconstruction: \url{https://en.wikipedia.org/wiki/Floyd-Warshall_algorithm\#Pseudocode_[11].}} move the datapoint $i$ that minimizes $s_{ij} - s_{ij'}$ from $j$ to $j'$. This will improve the assignment by incurring a negative total cost.
    \item Repeat until no negative cycle exists.
\end{itemize}
Depending on the initial assignment, only a small number of $O(k^3)$ improvements is needed and we find in practice for $k \ll n$ this algorithm is much faster than the $O(n^3)$ Hungarian \citep{kuhn1955hungarian} algorithm.

\subsection{Computing conditionally optimal assignments}
\label{app:compute_conditionally_optimal_assignments}
If the assignment is optimal, we denote the entries of the all-pairs shortest path matrix resulting from the Floyd-Warshall algorithm by $d^*_{jj'}$ (see Appendix \ref{app:gm_cycle_cancelling}). We can use this to efficiently obtain \emph{conditionally optimal assignments}, conditioning on $z_{ij} = 1$ for all $i, j$, as follows. If we condition on $z_{ij} = 1$, we may move datapoint $i$ from the globally optimal assignment $j^*$ to a suboptimal assignment $j$, incurring a cost $s_{ij^*} - s_{ij}$, and move another datapoint from $j$ to $j^*$ for a cost of $d^*_{jj^*}$, indirectly via the path found by the Floyd-Warshall algorithm. As such, the total cost of enforcing $z_{ij} = 1$ is $s_{ij^*} - s_{ij} + d^*_{jj^*}$. We denote the value of the globally optimal assignment by $v^*$. Subtracting the cost for enforcing $z_{ij} = 1$, we find that the value $v^*_{|z_{ij}=1}$ of the conditionally optimal assignment is given by
\begin{equation}
\label{eq:conditionally_optimal_fw}
    v^*_{|z_{ij}=1} = v^* - (s_{ij^*} - s_{ij} + d^*_{jj^*}) = v^* - s_{ij^*} + s_{ij} - d^*_{jj^*}.
\end{equation}

\subsection{Computation of the conditionals}
\label{app:gm_compute_conditionals}
The dependence of the Gumbel-Matching distribution on $\mathbf{x}$ is only through the logits $\mathbf{A} = (a_{ij})$, which allows us to slightly simplify notation in the rest of this section. Let $v^*(\mathbf{A}, \mathbf{G})$ be the value of the optimal assignment for the Gumbel-Matching problem with logits $\mathbf{A} = (a_{ij})$ and noise $\mathbf{G} = (g_{ij})$, and let $v^*_{|z_{ij}=1}(\mathbf{A}, \mathbf{G})$ be the value of the conditionally optimal assignment with the additional constraint that $z_{ij} = 1$ (see Appendix \ref{app:compute_conditionally_optimal_assignments}). Assuming a capacity $c = \frac{n}{k}$, then given that $z_{ij} = 1$, the problem reduces to assigning the remaining $n - 1$ datapoints to the remaining $(k - 1) \cdot \frac{n}{k} + (\frac{n}{k} - 1) = n - 1$ `slots' ($\frac{n}{k}$ for experts $j' \neq j$ and $\frac{n}{k} - 1$ for expert $j$). As this reduced problem does not depend on datapoint $i$, we denote with $\mathbf{A}_{-i}$ and $\mathbf{G}_{-i}$ the logits and Gumbels with row $i$ removed and we let $v^*_{-ij}(\mathbf{A}_{-i}, \mathbf{G}_{-i})$ be the value of the optimal assignment of the reduced problem. From the principle of optimality it follows that
\begin{equation}
\label{eq:gm_principle_of_optimality}
    v^*_{|z_{ij}=1}(\mathbf{A}, \mathbf{G}) = v^*_{-ij}(\mathbf{A}_{-i}, \mathbf{G}_{-i}) + a_{ij} / \tau + g_{ij}.
\end{equation}

We can use this to compute the desired conditionals. In a slight abuse of notation\footnote{$q(z_{ij})$ corresponds to $q(z_i)$ where $z_i = j$ but assumes a one-hot representation.} we write
\begin{align}
    & q(z_{ij}|\mathbf{A}, \mathbf{G}_{-i}) \notag \\
    =& P(z_{ij} = 1 | \mathbf{A}, \mathbf{G}_{-i}) \notag \\
    =& P\left(v^*_{|z_{ij}=1}(\mathbf{A}, \mathbf{G}) > \max_{j' \neq j} v^*_{|z_{ij'}=1}(\mathbf{A}, \mathbf{G})\middle| \mathbf{A}, \mathbf{G}_{-i}\right) \notag \\
    =& P\left(v^*_{-ij}(\mathbf{A}_{-i}, \mathbf{G}_{-i}) + a_{ij} / \tau + g_{ij} > \max_{j' \neq j} v^*_{-ij'}(\mathbf{A}_{-i}, \mathbf{G}_{-i}) + a_{ij'} / \tau + g_{ij'} \middle| \mathbf{A}, \mathbf{G}_{-i}\right) \notag \\
    =& \frac{\exp(v^*_{-ij}(\mathbf{A}_{-i}, \mathbf{G}_{-i}) + a_{ij} /\tau)}{\sum_{j'} \exp(v^*_{-ij'}(\mathbf{A}_{-i}, \mathbf{G}_{-i}) + a_{ij'} / \tau)} \notag \\
    =& \frac{\exp\left(v^*_{|z_{ij}=1}(\mathbf{A}, \mathbf{G}) - g_{ij}\right)}{\sum_{j'} \exp\left(v^*_{|z_{ij'}=1}(\mathbf{A}, \mathbf{G}) - g_{ij'}\right)} \label{eq:gm_q_cond_trick}
\end{align}
where we have used the Gumbel-max trick. Although $g_{ij}$ appears in \eqref{eq:gm_q_cond_trick}, $q(z_{ij}|\mathbf{A}, \mathbf{G}_{-i})$ does \emph{not} depend on $\mathbf{g}_i$ as this value cancels against $g_{ij}$ in \eqref{eq:gm_principle_of_optimality}. The values $v^*_{|z_{ij}=1}(\mathbf{A}, \mathbf{G})$ can be computed efficiently using Equation \eqref{eq:conditionally_optimal_fw} with the method described in Appendix \ref{app:compute_conditionally_optimal_assignments}.

\subsection{Unbiased of the Gumbel-Matching estimator}
Let $\mathbf{z} = \text{GM}(\log p(\cdot|\mathbf{x}),\mathbf{G})$ be the solution for the Gumbel-Matching problem with noise $\mathbf{G}$. The idea behind the Gumbel-Matching estimator is that we can derive an unbiased estimate of the gradient \emph{for each datapoint} as follows. First sample the Gumbel noise $\mathbf{G}_{-i}$ for all datapoints except $i$, and compute the conditional distribution $q_\theta(z_i|\mathbf{x},\mathbf{G}_{-i})$ (see Appendix \ref{app:gm_compute_conditionals} where $\mathbf{A} = \log p(\cdot|\mathbf{x})$ is the matrix with log-probabilities and $z_{ij} = 1 \Leftrightarrow z_i = j$). We can then use this conditional distribution over $z_i$ as a proposal distribution in \eqref{eq:reinforce}, which we can then reparameterize in terms of the Gumbel noise $\mathbf{g}_i$ for the $i$-th datapoint. Finally, we can use this estimator for all datapoints $i$, where we may reuse the same Gumbel noise and use their average as the estimate:
\label{app:gm_estimator}
\begin{align*}
    &\nabla \mathbb{E}_{\mathbf{z} \sim p_\theta(\mathbf{z}|\mathbf{x})}\left[\frac{1}{n} \sum_i f(x_i,z_i)\right] \\
    =& \frac{1}{n} \sum_i \mathbb{E}_{z_i \sim p_\theta(z_i|x_i)}\left[\nabla \log p_\theta(z_i|x_i)     f(x_i,z_i)\right] \\
    =& \frac{1}{n} \sum_i \mathbb{E}_{\mathbf{G}_{-i}}\left[\mathbb{E}_{z_i \sim q_\theta(z_i|\mathbf{x},\mathbf{G}_{-i})}\left[ \frac{p_\theta(z_i|x_i)}{q_\theta(z_i|\mathbf{x},\mathbf{G}_{-i})} \nabla \log p_\theta(z_i|x_i) f(x_i,z_i)\right]\right] \\
    =& \frac{1}{n} \sum_i \mathbb{E}_{\mathbf{G}_{-i}}\left[\mathbb{E}_{\mathbf{g}_i}\left[ \frac{p_\theta(z_i|x_i)}{q_\theta(z_i|\mathbf{x},\mathbf{G}_{-i})} \nabla \log p_\theta(z_i|x_i) f(x_i,z_i) \middle|_{z_i = \text{GM}(\log p(\cdot|\mathbf{x}),\mathbf{G})_i} \right]\right] \\
    =& \mathbb{E}_{\mathbf{G}}\left[\frac{1}{n} \sum_i \frac{\nabla p_\theta(z_i|x_i)}{q_\theta(z_i|\mathbf{x},\mathbf{G}_{-i})} f(x_i,z_i) \middle|_{\mathbf{z} = \text{GM}(\log p(\cdot|\mathbf{x}),\mathbf{G})}\right].
\end{align*}

\subsection{Maximum-entropy distribution over balanced assignments}
\label{app:maxent_distribution}
For simplicity, we assume $n = k$, but this can be easily generalized. Assume that we have a balanced (square) matrix (see Section \ref{sec:sinkhorn_balancing}) $\mathbf{P} = (p_{ij})$ with probabilities $p_{ij} \ge 0$ such that $\sum_i p_{ij} = \sum_j p_{ij} = 1$, i.e.\ the matrix $\mathbf{P}$ is \emph{doubly stochastic}. The Birkhoff decomposition \citep{birkhoff1946tres,von1953certain} decomposes such a matrix as a convex combination over \emph{permutation matrices} $\mathbf{Z} = (z_{ij})$, for which $z_{ij} \in \{0,1\}$ and $\sum_i z_{ij} = \sum_j z_{ij} = 1$. Such permutation matrices represent $n \times n$ matchings $\mathbf{z}$ in one-hot encoding (i.e.\ $z_{ij} = 1 \Leftrightarrow z_i = j$), which is why we will use $\mathbf{z}$ to denote them. A Birkhoff decomposition $\alpha_{\mathbf{z}} > 0, \sum_{\mathbf{z}} \alpha_{\mathbf{z}} = 1$ thus represents a joint probability distribution over matchings $\mathbf{z}$ (represented as permutation matrices) with marginal distributions (per datapoint) given by $\mathbf{P}$:
\begin{equation}
\label{eq:birkhoff_decomposition}
    p_{ij} = \sum_{\mathbf{z}} \alpha_{\mathbf{z}} z_{ij} \quad \forall i,j.
\end{equation}
Many such decompositions/distributions exist and can be found using the Birkhoff algorithm\citep{birkhoff1946tres}, but these will yield sparse $\alpha_{\mathbf{z}}$ and thus have low entropy and high dependence between the marginal distributions for different $i$. We aim to minimize this dependence between the marginal distributions, which is achieved by maximizing the entropy $- \sum_{\mathbf{z}} \alpha_{\mathbf{z}} \log \alpha_{\mathbf{z}}$ subject to the constraint \eqref{eq:birkhoff_decomposition} and $\sum_{\mathbf{z}} \alpha_{\mathbf{z}} = 1$, which has the Lagrangian:
\begin{equation*}
    \mathcal{L}(\bm{\alpha},\eta,\bm{\lambda}) = - \sum_{\mathbf{z}} \alpha_{\mathbf{z}} \log \alpha_{\mathbf{z}} - \eta\left(1 - \sum_{\mathbf{z}} \alpha_{\mathbf{z}}\right) - \sum_{ij} \lambda_{ij} \left(p_{ij} - \sum_{\mathbf{z}} \alpha_{\mathbf{z}} z_{ij}\right). 
\end{equation*}
This has first order conditions
\begin{align*}
    \frac{\partial}{\partial \alpha_{\mathbf{z}}} \mathcal{L}(\bm{\alpha},\bm{\lambda}) = - \log \alpha_{\mathbf{z}} - 1 + \eta + \sum_{ij} \lambda_{ij} z_{ij} = 0 & \quad \forall \mathbf{z} \\
    \frac{\partial}{\partial \lambda_{ij}} \mathcal{L}(\bm{\alpha},\bm{\lambda}) = p_{ij} - \sum_{\mathbf{z}} \alpha_{\mathbf{z}} z_{ij} = 0 & \quad \forall i,j \\
    \frac{\partial}{\partial \eta} \mathcal{L}(\bm{\alpha},\bm{\lambda}) = 1 - \sum_{\mathbf{z}} \alpha_{\mathbf{z}} = 0.
\end{align*}
If we let $u_{ij} = \exp(\lambda_{ij})$ and $w = \exp(\eta - 1)$, and convert from the `one-hot' representation $z_{ij}$ to the `indexing' representation $z_i$ (i.e.\ $\sum_{ij} \lambda_{ij} z_{ij} = \sum_i \lambda_{i,z_i}$), we find the solution
\begin{align*}
    \alpha_{\mathbf{z}} &= \exp \left(- 1 + \eta + \sum_{ij} \lambda_{ij} z_{ij}\right) = \exp(\eta - 1) \prod_{i} \exp(\lambda_{i,z_i}) = w \prod_{i} u_{i,z_i} \\
    p_{ij} &= \sum_{\mathbf{z}} \alpha_{\mathbf{z}} z_{ij} = \sum_{\mathbf{z}:z_i = j} \alpha_{\mathbf{z}} = w \sum_{\mathbf{z}:z_i = j} \prod_{i'} u_{i',z_{i'}} = w u_{ij} \sum_{\mathbf{z}:z_i = j} \prod_{i' \neq i} u_{i',z_{i'}} = w u_{ij} \cdot\text{Perm}(U_{-ij}) \\
    1 &= \sum_{\mathbf{z}} \alpha_{\mathbf{z}} = \sum_j \sum_{\mathbf{z}:z_i = j} \alpha_{\mathbf{z}} = \sum_j w u_{ij} \cdot \text{Perm}(U_{-ij}) = w \cdot \text{Perm}(U).
\end{align*}
Here $\text{Perm}(U)$ is the \emph{permanent} of the matrix $U = (u_{ij})$, and $U_{-ij}$ is the matrix $U$ with rows $i$ and $j$ removed. Since $w = 1 / \text{Perm}(U)$ we find
\begin{equation*}
    p_{ij} = \frac{u_{ij}\text{Perm}(U_{-ij})}{\text{Perm}(U)} \Rightarrow u_{ij} = \frac{p_{ij} \cdot \text{Perm}(U)}{\text{Perm}(U_{-ij})}
\end{equation*}
which can, in theory, be solved using a (very expensive) fixed point iteration scheme. Empirically we found that the solution takes the form $u_{ij} \approx p_{ij}^{1/\tau}$ for some $\tau$, such that the probability of an assignment $\mathbf{z}$ is given by
\begin{equation}
\label{eq:maximum_entropy_distribution}
    \alpha_{\mathbf{z}} = w \prod_{i} u_{i,z_i} = \frac{\prod_{i} u_{i,z_i}}{\text{Perm}(U)} \approx \frac{\prod_{i} p_{i,z_i}^{1/\tau}}{\text{Perm}(\mathbf{P}^{(1/\tau)})} \propto \left(\prod_{i} p_{i,z_i}\right)^{1/\tau} = \exp\left(\frac{1}{\tau} \sum_{i} \log p_{i,z_i}\right).
\end{equation}
Here $\mathbf{P}^{(1/\tau)}$ is the \emph{Hadamard power}, which raises the entries $p_{ij}$ to the power $1/\tau$ element-wise. By generalizing permanents (sums over permutations) to non-square matrices as sums over balanced assignment matrices, the same result can be derived for $n \neq k$. Using $a_{ij} = \log p_{ij}$ and converting to `one-hot' representation $z_{ij}$, we find that \eqref{eq:maximum_entropy_distribution} is equal to \eqref{eq:gibbs_distribution}.

If we do not use the approximation $u_{ij} \approx p_{ij}^{1/\tau}$, then it still holds that the maximum entropy distribution has the form \eqref{eq:gibbs_distribution}, but we should let $\tau = 1$ and $a_{ij} = \log u_{ij}$, so in this case $a_{ij} \neq \log p_{ij}$ are \emph{not} the marginal log-probabilities we aim to sample from.

\clearpage
\section{Experiment}
\label{app:toy_extra_results}
\begin{figure}
     \centering
     \begin{subfigure}[b]{0.32\textwidth}
         \centering
         \includegraphics[width=\textwidth]{images/reinforce_wide.pdf}
     \end{subfigure}
     \hfill
     \begin{subfigure}[b]{0.32\textwidth}
         \centering
         \includegraphics[width=\textwidth]{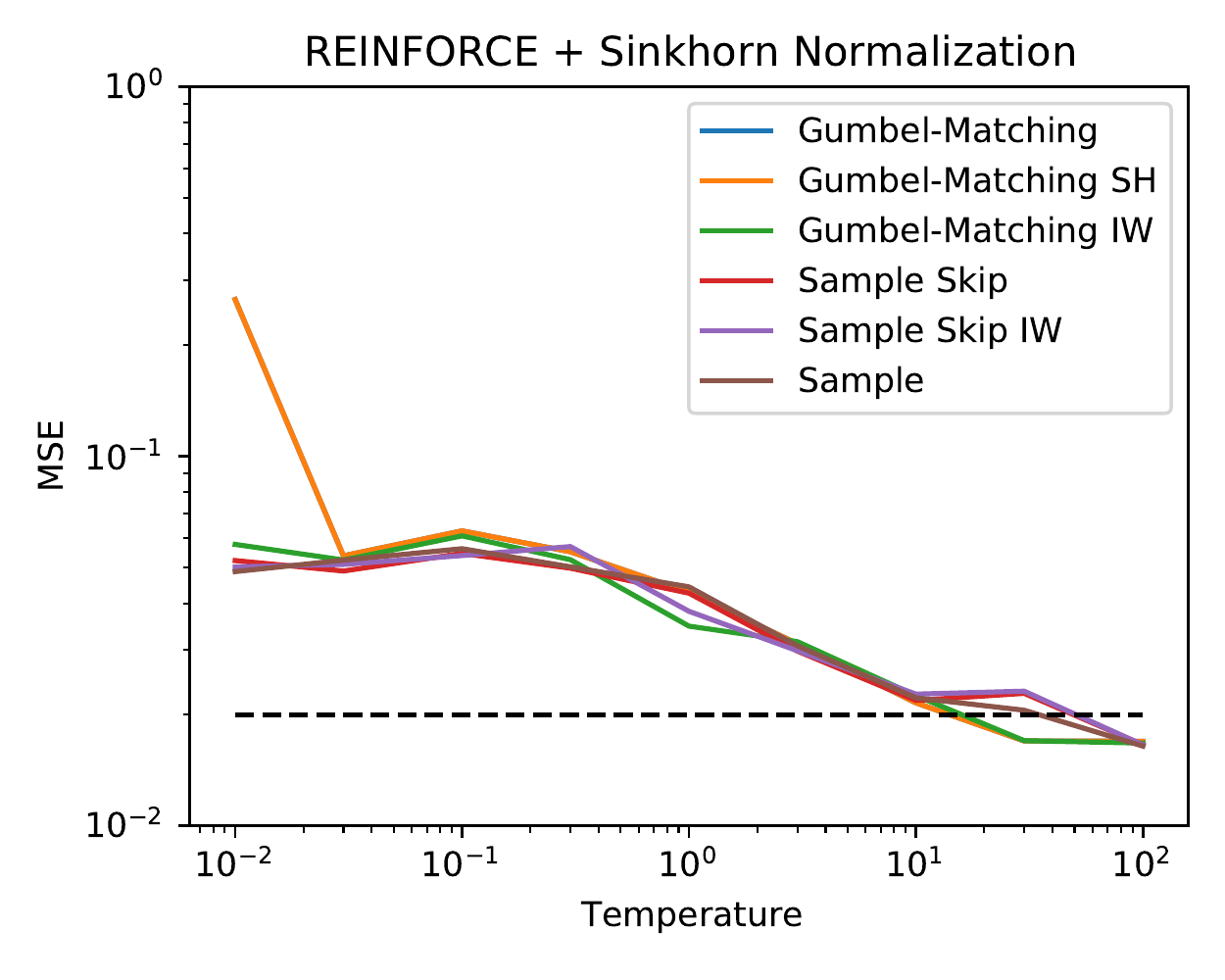}
     \end{subfigure}
     \hfill
     \begin{subfigure}[b]{0.32\textwidth}
         \centering
         \includegraphics[width=\textwidth]{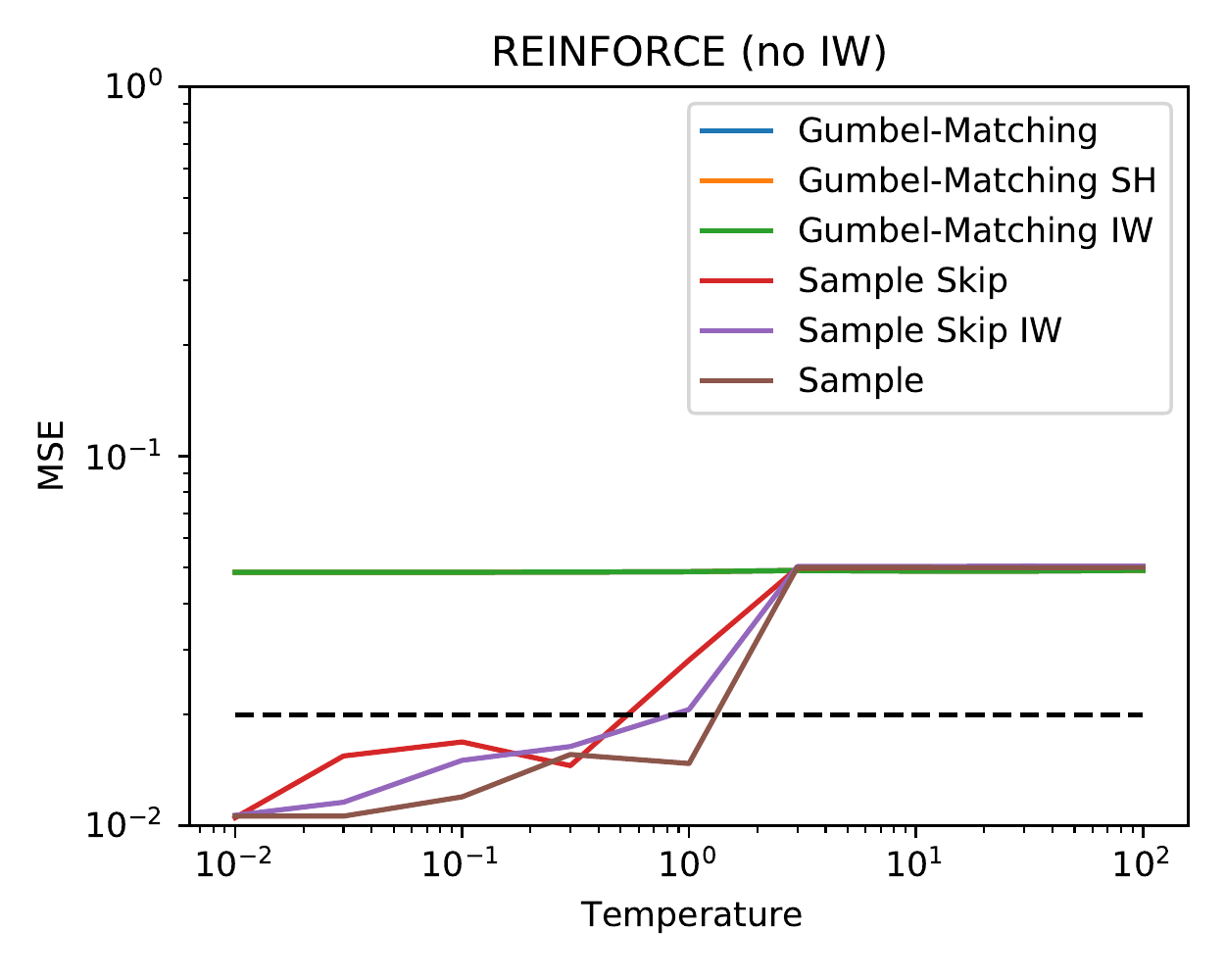}
     \end{subfigure}
     \\
     \begin{subfigure}[b]{0.32\textwidth}
         \centering
         \includegraphics[width=\textwidth]{images/gating_wide.pdf}
     \end{subfigure}
     \hfill
     \begin{subfigure}[b]{0.32\textwidth}
         \centering
         \includegraphics[width=\textwidth]{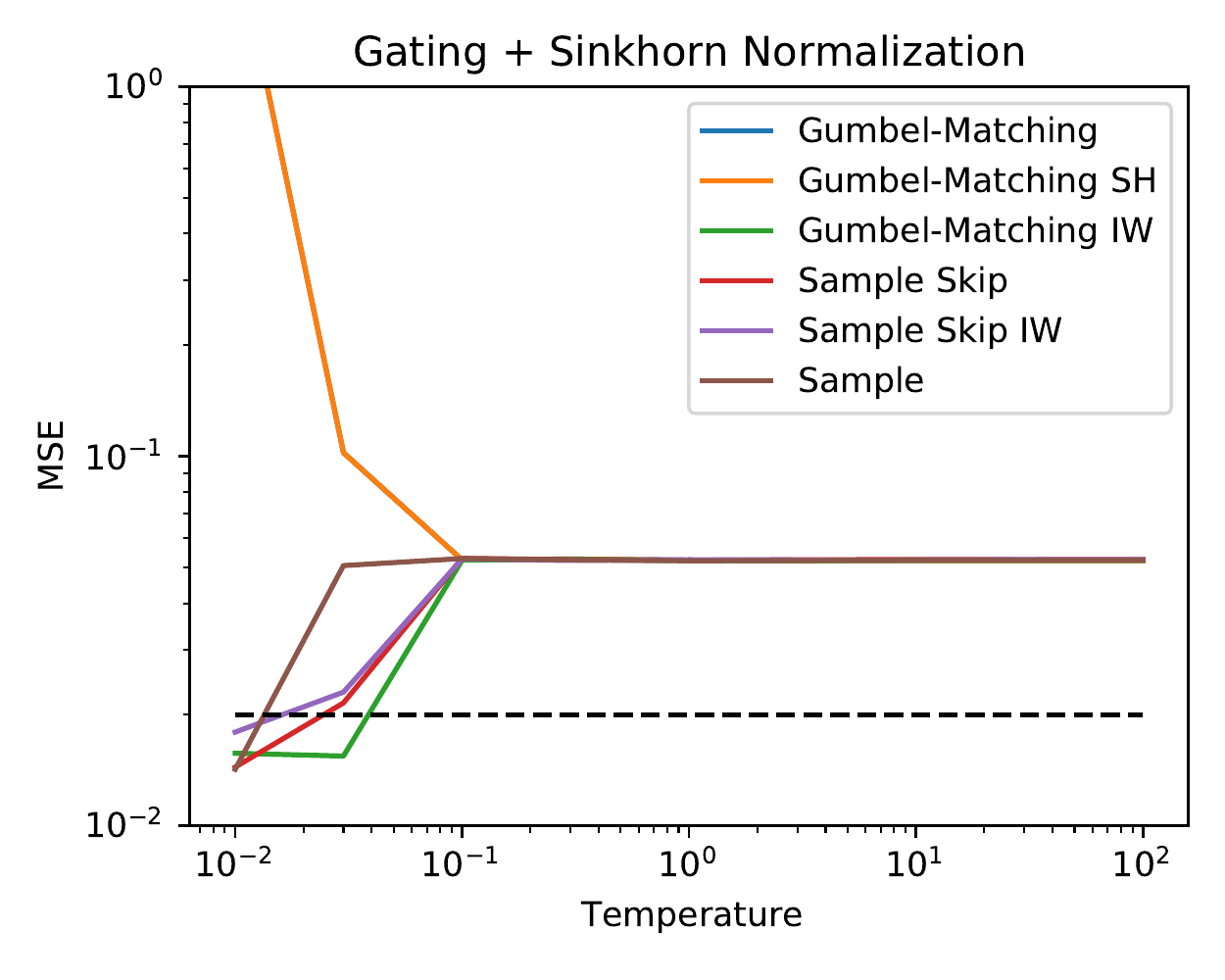}
     \end{subfigure}
     \hfill
     \begin{subfigure}[b]{0.32\textwidth}
         \centering
         \includegraphics[width=\textwidth]{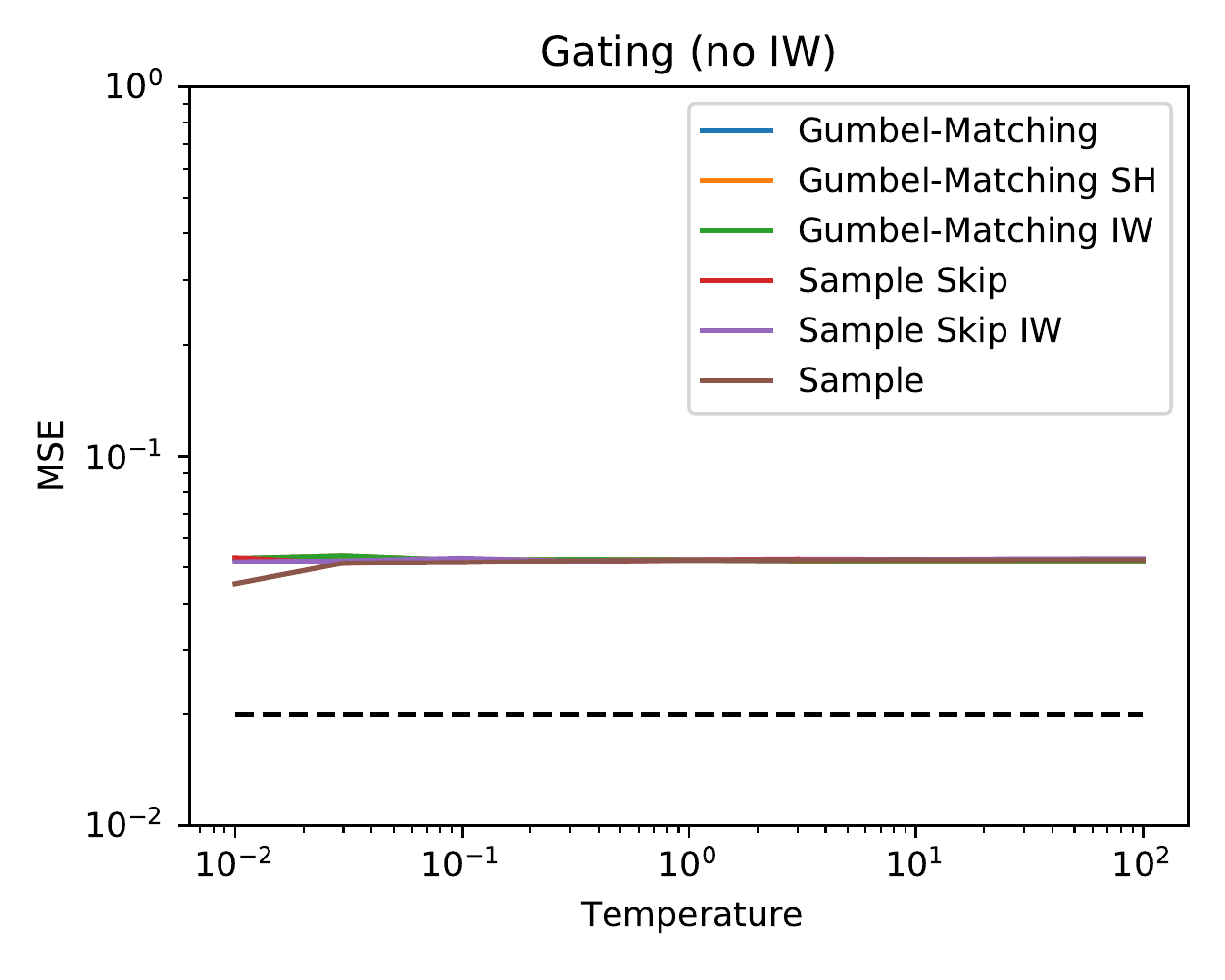}
     \end{subfigure}
     \caption{Results using REINFORCE (with no balance loss) and differentiable gating (with 0.01 balance loss weight) loss functions; both also shown in a version that applies Sinkhorn normalization before sampling, as well as a (biased) version that does not apply importance weights.}
     \label{fig:toy_results_appendix}
\end{figure}

Figure \ref{fig:toy_results_appendix} presents results for both REINFORCE (top) and differentiable gating (bottom) both with (middle) and without (left) the Sinkhorn normalization before sampling. When using Sinkhorn normalization before sampling, we take it into account when computing the importance weights. With REINFORCE, not using Sinkhorn normalization works better, as using Sinkhorn normalization requires a high sampling temperature to work well, i.e.\ close to uniform proposal samples. 

With differentiable gating (and load balancing loss with weight $0.01$), we observe the opposite: the results are better \emph{with} Sinkhorn normalization than without Sinkhorn normalization (Section \ref{sec:experiment_gating}), but, contrasting REINFORCE, require a very low sampling temperature to succeed, so close to deterministic training. 

Existing methods often do not use importance weights to take into account the noise scale or temperature \citep{shazeer2017outrageously,fedus2021switch}, so we experiment with this as well by completely dropping importance weights for all estimators (right column in Figure \ref{fig:toy_results_appendix}). We find this only works for REINFORCE, when sampling with a low temperature but not when we use the Gumbel-Matching estimators.

In all cases, we found REINFORCE succeeds both with and without a load balancing loss, whereas differentiable gating requires a load balancing loss with a weight of 0.01 to succeed at all. If the load balance loss is too high (0.1) all methods fail to model the unbalanced data.

\end{document}